\documentclass[preprint,12pt]{elsarticle}

\usepackage{amssymb}
\usepackage{amsmath,amsfonts}
\usepackage[ruled]{algorithm2e}
\usepackage{array}
\usepackage[caption=false,font=normalsize,labelfont=sf,textfont=sf]{subfig}
\usepackage{textcomp}
\usepackage{stfloats}
\usepackage{url}
\usepackage{verbatim}
\usepackage{graphicx}
\usepackage{multirow,tabularx}
\usepackage[scaled]{beramono}
\usepackage{booktabs}
\newtheorem{definition}{Definition}
\usepackage[table]{xcolor}
\usepackage{multirow}
\usepackage[raggedright]{sidecap}
\usepackage{makecell}
\usepackage{multirow,tabularx}
\usepackage{longtable}
\usepackage{booktabs}
\usepackage [english]{babel}
\usepackage [autostyle, english = american]{csquotes}
\MakeOuterQuote{"}

\journal{Information Sciences}

\begin{document}

\begin{frontmatter}
\title{Anomaly Detection and Classification in Knowledge Graphs}

\author{Asara Senaratne, Peter Christen, Pouya Omran, Graham Williams \\ 
{\texttt{\{asara.senaratne, peter.christen, p.g.omran, graham.williams\}@anu.edu.au} \\}
School of Computing \\ Australian National University.}

\begin{abstract}
Anomalies such as redundant, inconsistent, contradictory, and deficient values in a Knowledge Graph (KG) are unavoidable, as these graphs are often curated
manually, or extracted using machine learning and natural language processing techniques. Therefore, anomaly detection is a task that can enhance the quality of KGs. In this paper, we propose SEKA (\underline{Se}eking \underline{K}nowledge Graph \underline{A}nomalies), an unsupervised approach for the detection of abnormal triples and entities in KGs. SEKA can help
improve the correctness of a KG whilst retaining its coverage.  We propose an adaption of the Path Rank Algorithm (PRA), named the Corroborative Path Rank Algorithm (CPRA), which is an efficient adaptation of PRA that is customized to detect anomalies in KGs. Furthermore, we also present TAXO (\underline{Taxo}nomy of anomaly types
in KGs), a taxonomy of possible anomaly types that can occur in a KG. This taxonomy provides a classification of the anomalies discovered by SEKA with an extensive discussion of possible data quality issues in a KG. We evaluate both approaches using the four real-world KGs YAGO-1, KBpedia, Wikidata, and DSKG to demonstrate the ability of SEKA and TAXO to
outperform the baselines.
\end{abstract}




\end{frontmatter}

\section{Introduction}
\label{sec:introduction}
KGs form the backbone of many knowledge-dependent
applications such as search engines and digital personal assistants.
When constructing a KG, it can either be
manually curated by experts, manually generated by volunteers, automatically extracted from text via hand-crafted or learned
rules, or automatically extracted from unstructured text using 
ML techniques. Hence, it is unrealistic
to expect a perfect archive of knowledge. While validation techniques such as Shapes Constraint
Language (SHACL)\footnote{\label{fn1}\url{https://www.w3.org/TR/shacl/}} and Shape Expressions (ShEx)\footnote{\label{fn2}\url{https://shex.io/}} language offer insights
into the structure of a KG~\cite{delva2023shacl, rabbani2022shaclshex},
not every real-world KG has a shape-based layer to facilitate such
validation. Furthermore, these techniques propose what should be
in a KG as opposed to what should not be in a KG. 

Similarly, rule-based reasoners and constraint engines for KG 
validation~\cite{huaman2021weightedkgvalidation} only find common patterns of errors.
Even though errors can be represented via pre-known patterns~\cite{dong2023activelearning}, an
anomaly cannot be guessed before being detected. While an
error can be considered as an anomaly in data, not every anomaly
is erroneous. Non-erroneous anomalies can potentially
uncover interesting information, thus discovering new knowledge
from a KG. Although there exist other approaches to detect anomalies in
KGs, they are either domain-specific~\cite{wienand2014dbpediaanomalies}, 
require human involvement, or are dependent on external resources~\cite{huaman2021weightedkgvalidation}.

Hence, in this paper, we propose SEKA (\underline{See}king \underline{K}G \underline{A}nomalies), an unsupervised anomaly detection pipeline to
identify anomalous triples and entities in a KG using both
structural properties and content of the graph. To the best of
our knowledge, there is no other work in the domain of KG
validation that can detect both anomalous facts and entities in a KG. Our aim is 
to discover abnormal triples and entities that provide 
interesting, unusual, contradicting, semantically 
incorrect, redundant, invalid, incomplete, and missing 
information in KGs, as provided with examples in 
Table~\ref{tab:anomalytypes}. Furthermore, to infer missing type information of entities using named entity recognition~\cite{nadeau2007ner}, we introduce ENTGENE (\underline{En}tity \underline{T}ype \underline{Gene}rator). Following general KG terminology~\cite{ehrlinger2016towardskg}, we use the term \emph{entity} to 
refer to a node in the KG that represents a real-world object 
(such as a person, movie, dataset, and so on), and we 
interchangeably use the terms \emph{fact} and \emph{triple} to 
represent two related entities. 

\begin{table*}[!th]
\caption[Examples of interesting anomalies detected by SEKA.]{Examples of interesting anomalies detected by SEKA in YAGO-1, DSKG, Wikidata, and KBpedia.}
\begin{center}
\scalebox{0.8}{
\begin{tabular}{p{3in}p{3.5in}}
    \toprule
    Anomalous triple & Anomaly explained\\
    \midrule 

    (DonaldTrump, marriedTo, MarlaMaples) & \multirow{2}{\linewidth}{Two contradicting relationships between the same two people, while one triple is wrong.}\\
    (MarlaMaples, hasChild, DonaldTrump) & \\
    
    \addlinespace[2pt]
    \hline 
    \addlinespace[2pt]

   \multirow{2}{\linewidth}{{(EthelricArchbishopOfYork, hasSuccessor, AelfricPuttoc) \newline (EthelricArchbishopOfYork, hasPredecessor, AelfricPuttoc)}} &
    It is unusual for one person to be both the predecessor and successor of another person. However, this can be a possibility in politics and religion.\\
    
    \addlinespace[15pt]
    \hline 
    \addlinespace[2pt] 
    
    \multirow{3}{\linewidth}{(KarlHermannKnoblauch, hasWonPrize, KingdomOfPrussia)} &
    The predicate "hasWonPrize" is generally followed by the name of the prize won instead of the object for which it was awarded, thus making the predicate usage ambiguous.\\

    \addlinespace[2pt]
    \hline 
    \addlinespace[2pt] 
    
    (Ain'tTooProudToBeg, isOfGenre, rock) & \multirow{3}{\linewidth}{The range of this predicate is not well defined. Hence, the same subject and predicate have different objects.}\\
    (Ain'tTooProudToBeg, isOfGenre, music) & \\
    (Ain'tTooProudToBeg, isOfGenre, popularMusic) & \\

    \addlinespace[2pt]
    \hline 
    \addlinespace[2pt] 

    (AMGrapper, produced, BettaHaveMoney) &
    \multirow{2}{\linewidth}{These two facts seem to provide redundant information causing entity redundancy.}\\
    (AMGrapper, produced, BettaHaveMoney2001)\\

    \addlinespace[2pt]
    \hline 
    \addlinespace[2pt] 
    
    \multirow{2}{*}{(Plato, bornOn, "Athens")} & The object contains an invalid value even though Athens is where he was born.\\

    \addlinespace[2pt]
    \hline 
    \addlinespace[2pt] 
    
    (Marcelona, bornIn, Mozambique) & \multirow{2}{\linewidth}{Entity "Mozambique" is treated both as a person and location causing entity ambiguity.}\\
    (Marcelona, hasSuccessor, Mozambique)\\

    \addlinespace[2pt]
    \hline 
    \addlinespace[2pt] 
    
    \multirow{2}{\linewidth}{(9thWonder, produced, TheDreamMerchantVol2)} & The subject is missing its corresponding "created" triple which other triples related to music albums have, thus making the entity rare.\\

    \addlinespace[2pt]
    \hline 
    \addlinespace[2pt] 
    
    (Neuromance, isOfGenre, Hacker) & "Hacker" is not a common type of genre. Hence the object seems abnormal.\\

    \addlinespace[2pt]
    \hline 
    \addlinespace[2pt] 
    
    (Person/11203, hasName, A.) & "A." alone cannot be the name of a person, thus creating an abnormal object\\

    \addlinespace[2pt]
    \hline 
    \addlinespace[2pt] 
    
    (SQL, hasDefinition, "") & A triple with a missing literal value.\\

    \addlinespace[2pt]
    \hline 
    \addlinespace[2pt] 
    
    \multirow{2}{*}{(Dataset/410)} & This is an anomalous entity as it is the only dataset with eleven creators, whereas other datasets have at most five creators.\\

    \bottomrule
\end{tabular}}
\label{tab:anomalytypes}
\end{center}
\end{table*}

\pagebreak
Having detected anomalies, we must obtain a classification among these identified anomalies such that we know what anomalies to be forwarded to domain experts for correction, and what can be corrected via automatic or semi-automatic techniques. However, to the best of our knowledge, in the literature, there is no such pre-defined classification of possible common anomalies that could arise in a KG, which we could directly use to support anomaly classification. As there is no such taxonomy of anomaly types introduced for KGs, we next propose TAXO (\underline{Taxo}nomy of anomaly types
in KGs), a unification of an extensive set of anomaly types in KGs that we can discover either by
analyzing KG data storage files such as Notation3 (N3)\footnote{\url{https://www.w3.org/TeamSubmission/n3/}}, or by representing data as a graph. 

TAXO can support domain experts to prevent the identified anomalies from occurring in new KGs, and to broaden their view when developing algorithms to identify anomalies in existing KGs. The ultimate contribution of our work is towards
the enhancement of data quality thereby generating enriched KGs. A unifying view of anomalies provides a solid foundation to understand the severity of these anomalies, discovers knowledge, and supports future research in this area.

Next, we present a review of the existing literature in the domain of anomaly detection. Subsequently, we outline the preliminaries required to discuss the approaches SEKA, ENTGENE, and TAXO in detail. Following this, we present the computational complexity of the proposed solutions and their experimental evaluation, before concluding the paper.

\section{Related Work}
\label{sec:relatedwork}
Anomaly detection in KGs has received much attention as automated
methods of constructing KGs prioritize data
integrity~\cite{heindorf2016wikivandalism}. There are a few methods
for error detection in KGs, where each approach may target specific
types of information~\cite{paulheim2017kgrefinement}. There are also approaches taking advantage of entity type information to perform clustering-based outlier
detection~\cite{debattista2016improvelinkeddata}. However, entity
types can either be absent or only partially available in real-world KGs.
Alternatively, another group of methods uses path
ranking~\cite{melo2017detectrelationassertions}, and path-based rule
mining for error checking~\cite{shi2016factchecking}. While path-ranking
methods have limitations in the coverage, path-based rule mining
methods have limitations in the quality of the rules.

Most of the
recently conducted studies propose to employ supervised classification to evaluate every triple-based
on different features, including entity categories, path features,
in/out-degrees, as well as embedding representations of entities and
relations~\cite{melo2017detectrelationassertions}. However, ground truth data may not be available to train such classifiers~\cite{vu2024metalearning}.
Alternatively, there have been efforts to utilize external
information
sources~\cite{wang2020kgvalidation}, such as related web
pages~\cite{lehmann2012factvalidation} and
annotations~\cite{liu2021errordetection},
to facilitate anomaly detection. While having external resources can
be valuable for this task, acquiring such supplemental information
is time-consuming and expensive. 

Furthermore, there are approaches
that only aim to identify a single type of
anomaly~\cite{melo2017detectrelationassertions}, approaches that are
KG
dependent~\cite{paulheim2017dbpedia},
methods requiring human
intervention~\cite{jeyaraj2019kgerrordetection}, and embedding
methods that can only consider structural properties eliminating
the content associated with entities~\cite{abedini2020correctiontower}.

To validate KGs by offering insights into KG
structure, techniques such as Shapes Constraint Language
(SHACL)\footref{fn1} and Shape Expressions
(ShEx)\footref{fn2} language~\cite{rabbani2022shaclshex} have also been introduced. Given these techniques that aim to enhance KG quality, it is important to have a common language to
describe anomalies, as provided by a taxonomy. In addition to the need for a common language to describe anomalies, the ability to support anomaly detection and correction techniques is opening avenues for future research facilitating the creation of  taxonomies~\cite{wiseman2011errortaxonomy}. 

Taxonomies are important to better understand issues, challenges, and trends in various 
domains~\cite{butt2015taxonomy}. For example, in the domain of the semantic web, the work by Breit et al.~\cite{breit2023mltaxonomy} provides a classification for machine learning-based semantic web systems which can be used as a template to analyze existing semantic web systems and to describe new ones. Even though the importance of taxonomies has been
identified decades ago~\cite{clement1985misconceptionstaxonomy}, to the best of our 
knowledge, a taxonomy of anomaly types in KGs has not been proposed so far. We therefore propose a taxonomy of anomaly types in KGs in Section~\ref{sec:taxo}.

\section{Preliminaries}
\label{sec:preliminaries}

Considering an edge-labelled graph,  which is a type of an attributed graph with a single categorical attribute (label) for the edge~\cite{Wang2021attributedgraphs}, we define a \emph{path} in such a graph as follows:

\begin{definition}[Path]
    In an edge-labelled graph $G$, a path $\mathcal{P}$ is defined as a directed, labelled sequence of vertices and edges $v_1 \xrightarrow[]{\text{p}_1} v_2 \xrightarrow[]{\text{p}_2} ... \xrightarrow[]{\text{p}_{k-1}} v_k$  in $G$, where $v_i \in V$ denotes real-world entities, $\text{p}_i$ represents the predicate (edge label) of the directed edge that connects vertex $i$ to $i+1$, and $k$ denotes the length of the path.
\end{definition}

Considering a \textit{directed} edge-labeled graph as defined above, we now define
the \emph{neighborhood} of a vertex as follows~\cite{Wang2021attributedgraphs}:

\begin{definition}[Neighborhood]
    For an edge-labeled graph $G$, the neighborhood $N_G(v)$ of a 
    vertex $v \in V$ is the set of all neighbors of $v$, $N_G(v) = \{u | \{u, v\} \in E\}$; $u \in V$. 
\end{definition}

\noindent Finally, we define a KG which is a directed edge-labeled 
graph~\cite{Wang2021attributedgraphs} as follows:

\begin{definition}[Knowledge Graph]
We consider a directed edge-labeled KG, $G = (V, E)$ containing a
set of nodes (or vertices) $V$ and a set of labeled edges $E$
connecting these vertices. 
The Resource Description Framework (RDF)\footnote{\url{https://www.w3.org/TR/rdf-concepts/}} is a standardized data
model based on directed edge-labeled
graphs. The RDF model defines three types of nodes in a KG:
(1) Internationalized Resource Identifiers (IRIs) which assign
a global identifier for entities (the set of entities with an IRI $\mathbf{I}_e$) and relations (the set of relations with an IRI $\mathbf{I}_r$) on the
web (where $\mathbf{I} = \mathbf{I}_e \cup \mathbf{I}_r$); (2) literals which represent strings and other datatype values (set of literals $\mathbf{L}$); and (3) blank nodes which
are anonymous nodes (not an IRI reference or a literal) that do not
have an identifier (the set of blank nodes $\mathbf{B}$)~\cite{hogan2021kg}. We therefore have the node
set $V = (\mathbf{I}_e \cup \mathbf{L} \cup \mathbf{B})$, and edge set $E \in V \times \mathbf{I}_r
\times V$. Each edge is considered as an RDF triple (triplet) or a statement of fact $F = (s,p,o)$, where
subject $s \in S$, predicate $p \in P$, object $o \in O$, $s \xrightarrow[]{\text{p}} o$, $(s,o) \in V$, and ($S \times P \times O) \in E$. Furthermore, $s \in (\mathbf{I}_e$ or $\mathbf{B})$, $p \in \mathbf{I}_r$, and $o \in (\mathbf{I}_e$ or $\mathbf{L}$ or $\mathbf{B})$.
\end{definition}

An entity or a fact is classified \textit{abnormal}, if the associated data deviates significantly from the rest of the data under consideration.

\section{SEKA: Seeking Knowledge Graph Anomalies}
\label{sec:seka}

Following similar work in anomaly detection in
KGs~\cite{felfernig2012ckb, jia2018patterndiscoverykg, mitropoulou2024anomalydetection, wienand2014dbpediaanomalies}, we aim to discover abnormal triples and entities,
on the basis that they are rare,
missing, inconsistent, duplicate, incomplete, or interesting
patterns in the context of a given KG. 

\pagebreak
For example, an abnormal
triple can be one that has a relationship between two entities causing a contradiction with another relationship of the same two entities. We view the anomaly detection problem as an unsupervised
learning task that validates a proposed triple or an entity by determining if the data associated with the triple and entity can be verified using the data within the KG. To ensure high data quality, and to extract 
accurate insights from data, it is critical that such abnormalities are detected so they can be investigated.

\begin{figure*}[t!]
  \centering
  \includegraphics[width=1\textwidth]{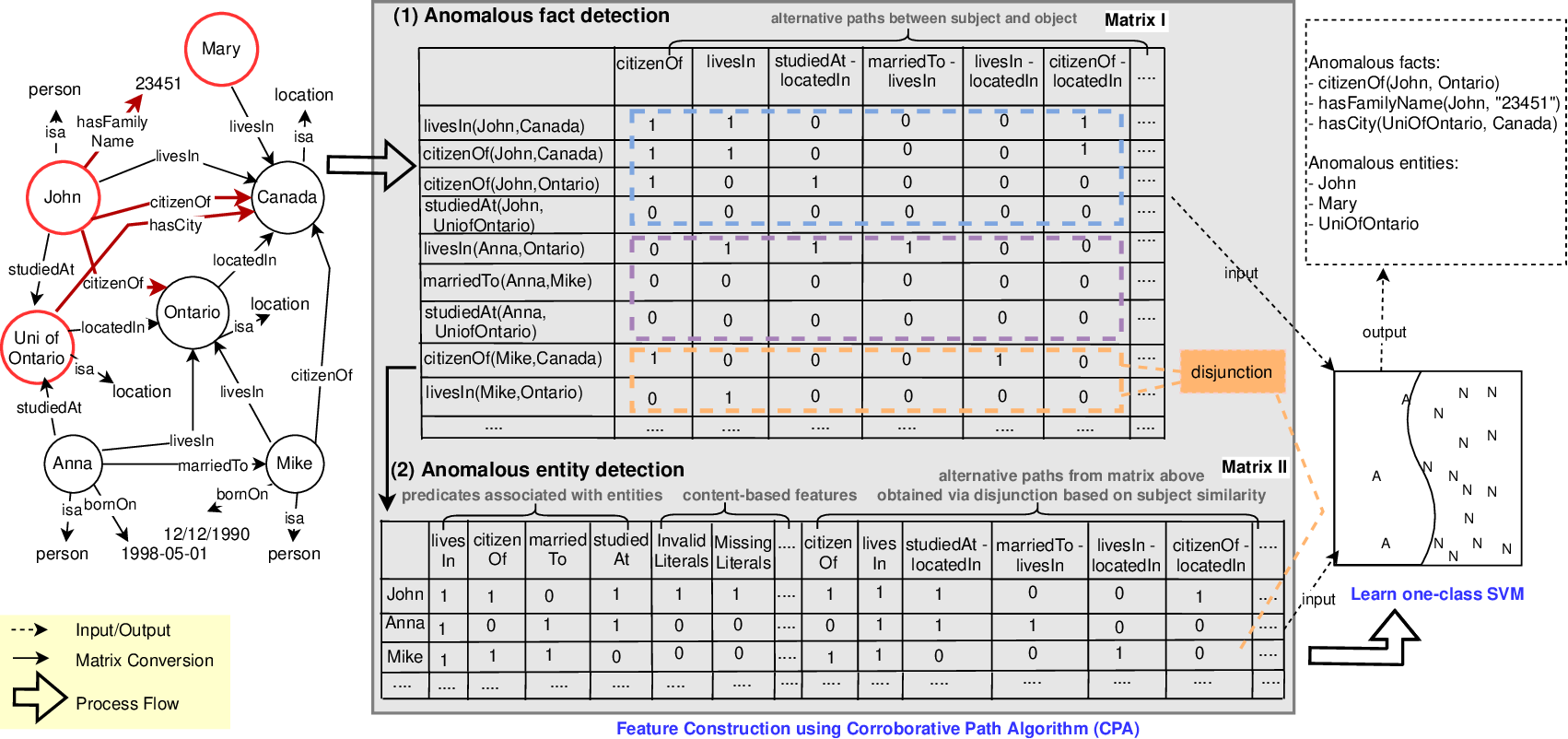}
  \caption[Overview of SEKA.]{Overview of SEKA, the anomaly detection process to
  identify anomalous triples and entities in a KG. The abnormal triples and entities are marked in red in the KG on the left-hand side of the figure (taken from~\cite{senaratne2023seka}).}
  \label{fig:seka}
\end{figure*}

Our path-based approach, the Corroborative Path Algorithm (CPA) which we introduce as shown in Figure~\ref{fig:seka}, performs the two tasks; fact anomaly detection to identify anomalous triples, and entity
anomaly detection to identify anomalous entities in a KG. We visualize the first task in the first (top) matrix (Matrix~I) in Figure~\ref{fig:seka}. In this
matrix (Matrix~I), a row represents a triple from the KG on the left of Figure~\ref{fig:seka},
the features are the alternative paths between entities with a path
length of up to two. These features are binary and indicate the 
existence or non-existence of a path between two entities. For
example, the two entities \emph{John} and \emph{Canada} in the
triple  \emph{livesIn(John, Canada)} have the alternative paths
\emph{citizenOf} and
\emph{citizenOf---locatedIn} as 
indicated by the binary value 1 (true). 

\pagebreak
The second task is to identify anomalous entities, where we
identify abnormal entities considering both structural
properties~\cite{Henderson2011KDD} and content associated with an entity, as shown by the
second (bottom) matrix (Matrix~II) in Figure~\ref{fig:seka}. The aim of this task is to identify entities that can be
anomalous even when there are no anomalous facts associated with
them. For example, consider the node \emph{Mary} in the KG of Figure~\ref{fig:seka}, which is abnormal due to the absence of associations compared to other nodes in the KG. In the second matrix, a row represents an entity from the KG
on the left of Figure~\ref{fig:seka}, while the columns represent three types of features. The
first set of features (structural) indicates the predicates the
entity is associated with within its neighborhood. The second set
of features (content-based) highlights data quality aspects by
referring to the literal-based triples associated with an entity. We obtain the third set of
features (structural) via the disjunction of the feature vectors
from Matrix~I, where the entity is the subject of the triple. We can
then either use Matrix~I or Matrix~II as input to a one-class $\nu$-SVM
for unsupervised learning to obtain abnormal triples or entities, respectively.

\subsection{Feature Generation}

Based on a set of feature generation functions $\pmb{\mathcal{F}}$, where
$\pmb{\mathcal{F}}_s$ represents structural features and
$\pmb{\mathcal{F}}_c$ content-based features, and $\pmb{\mathcal{F}} =
\pmb{\mathcal{F}}_s \cup
\pmb{\mathcal{F}}_c $, we generate the two feature matrices
$\mathbf{F}_x$ (Matrix~I in Figure~\ref{fig:seka}) for fact anomaly detection and $\mathbf{F}_y$ (Matrix~II in Figure~\ref{fig:seka}) for entity
anomaly detection, such that
$\pmb{\mathcal{F}}_s$ determines the feature matrix $\mathbf{F}_t$, and
both $\pmb{\mathcal{F}}_s$ and $\pmb{\mathcal{F}}_c$ together determine the feature matrix $\mathbf{F}_e$. To construct features that highlight the structural properties of
a KG, we introduce a novel variation of the Path Rank Algorithm (PRA)~\cite{lao2011pra}, 
named the Corroborative Path Algorithm
(CPA). While PRA is widely used for the task of link
prediction~\cite{lao2011pra}, CPA is dedicated for anomaly detection. The following are the characteristics of CPA.
\begin{itemize}
    \item CPA considers \textbf{all} paths $\pmb{\mathcal{P}}$ between two entities ($s\in
    S $ and $o \in O$, where $ (s,o) \in \mathbf{I}_e$) up-to a given path
    length $k$. To reduce complexity and for ease of interpretation, we bound $k$ as we discuss below.
    Referring to the KG in
    Figure~\ref{fig:seka}, between the two entities \emph{John}
    ($s \in S$) and \emph{Canada} ($o \in O$), there exists three paths,
    where \emph{(livesIn)} and \emph{(citizenOf)} are of $k=1$ as given in
    expression~\ref{eq:lengthone} below, and \emph{(citizenOf, locatedIn)} is of $k=2$ as
    given in expression~\ref{eq:lengthtwo}. Here, we also consider the inverse predicate $p_1(s,o) \wedge p_2(o,s)$. 

    \begin{eqnarray}
    p_1(s,o) \wedge p_1(s,w), w \neq o \label{eq:lengthhalf1}\\
    p_1(s,o) \wedge p_2(s,w), w \neq o \label{eq:lengthhalf2} \\
    p_1(s,o) \wedge p_3(s,o) \label{eq:lengthone} \\
    p_1(s,o) \wedge p_4(s,z) \wedge p_5(z,o), z (\neq s,o)  \label{eq:lengthtwo}
    \end{eqnarray}
    
    We introduce the
    concept of \emph{half path} (paths of length $k=0.5$) to determine the
    occurrence of a given subject $s \in \mathbf{I}_e$ and predicate $p \in P$ together with any other object $o \in \mathbf{I}_e$ as opposed to the object under
    consideration. That is, we consider half of a triple, which is the
    subject and predicate to determine its other occurrences. For example, \emph{John} holds the relationship \emph{citizenOf} with the two entities \emph{Canada} and \emph{Ontario}. We demonstrate this in expression~\ref{eq:lengthhalf1}. We consider these paths as features in $\mathbf{F}_x$. 

    Another half path we consider when constructing the entity features for $\mathbf{F}_y$ is the occurrence of a given subject $s \in \mathbf{I}_e$ in a particular triple with any other object $o \in \mathbf{I}_e$ and predicate $p \in P$ as opposed to the object and predicate under consideration, as demonstrated in expression~\ref{eq:lengthhalf2}. In expression~\ref{eq:lengthhalf1}, \ref{eq:lengthhalf2}, and \ref{eq:lengthtwo}, $w$ and $z$ represent free variables that are quantified existentially.
    
    \item While the existing PRA performs random walks over the graph, CPA performs a depth-first search~\cite{sedgewick2001algorithms} from the starting node to identify $\mathcal{P} \in \pmb{\mathcal{P}}$ between a given $s \in \mathbf{I}_e$
    and $o \in \mathbf{I}_e$. CPA does not perform random walks as we are not interested in calculating the probability of arriving at a particular object starting from a particular subject, given a random walk exactly following all the relations between them~\cite{wang2016pra}. Since our aim is to identify anomalous triples and entities using the paths that connect these entities, we use \emph{all} paths between $(s,o) \in \mathbf{I}_e$ bounded by $k$, instead of a probabilistic approach. Furthermore, generating binary features improves the explainability of the anomalies. For example, the triple \emph{(Mary, livesIn, Canada)} is abnormal due to the path \emph{originatedIn} that exists between
    \emph{Maya} and \emph{Canada}.

    \pagebreak
    \item The advantage of CPA is that it has lower complexity and substantially lower run times compared to traditional PRA, making CPA scalable and well-suited for anomaly detection in large KGs (we demonstrate this in Section~\ref{subsec:experimentsseka}). Furthermore, with the generation of semantic features, CPA has the capability of detecting semantic anomalies as we consider the sequence of 
    occurrence of paths between two entities as the features, which have the 
    potential to identify rare path occurrences such as
    \emph{marriedTo---hasChild}, which is an example from Table~\ref{tab:anomalytypes}. 

\end{itemize}

\begin{algorithm*}[t!]
\scriptsize
\caption{Generation of structural features}
\label{alg:structuralfeatures}
\SetAlgoNoLine
\KwIn{$G$: \hspace{1.4mm} The knowledge graph.}
\KwOut{$\mathbf{F}_x$: \hspace{0.1mm} The matrix for fact anomaly detection.}
\smallskip
\begin{tabbing}
0:~ \hspace{3mm} $\mathbf{C} \leftarrow
    setfeaturesfornodesfromattributetypeandkindofany(G, F, L)$~\= \kill
\hspace{3mm} \tiny{// Get triples from $G$, where $s$ and $o$ are entities} \\
1:~ $G_e \leftarrow getTriples(G)$ \> \\

\hspace{3mm} \tiny{// Maximum path length to traverse}\\
2:~ $k \leftarrow getMaxPathLen(default={0.5,1,2})$ \>  \\       

\hspace{3mm} \tiny{// Iterate over each triple in $G_e$} \\
3:~ \textbf{for} $triple$ in $G_e$: \>\\

 \hspace{9mm}\tiny{// Get $s$ and $o$ of a triple} \\
4:~ \hspace{3mm} $s,o \leftarrow getSubject(triple), getObject(triple)$ \>\\

\hspace{8mm} \tiny{// Get all paths between $s$ and $o$ bounded by $k$} \\
5:~ \hspace{3mm} $pathList \leftarrow getAllPaths(s, o, k, G_e)$ \>\\

\hspace{8mm} \tiny{// Aggregate paths of every triple to a dictionary, where $key=triple$, and $value=pathList$} \\
6:~ \hspace{3mm} $dictOfPaths \leftarrow concatPathsToList(dictOfPaths, triple, pathList)$ \>\\

\hspace{3mm} \tiny{// Get unique list of paths from $dictOfPaths$}\\
7:~ $uniquePathsList \leftarrow getUniquePaths(dictOfPaths.values)$\>\\

\hspace{3mm} \tiny{// Build matrix with unique paths as features}\\
8:~ $\mathbf{F}_t \leftarrow constructFeatureMatrix(uniquePathsList)$\>\\

\hspace{3mm} \tiny{// Iterate over all triples}\\
9:~ \textbf{for} $triple$ in $dictOfPaths.keys$: \>\\

\hspace{8mm} \tiny{// Iterate over all unique features in $\mathbf{F}_t$}\\
10:~ \hspace{3mm}\textbf{for} $feature$ in $uniquePathsList$: \>\\

\hspace{12mm} \tiny{// Check if triple has that particular feature as path}\\
11:~ \hspace{6mm} $binaryVal \leftarrow tripleHasFeature(dictOfPaths, triple, feature)$ \>\\

\hspace{12mm} \tiny{// Construct the feature vector of the triple}\\
12:~ \hspace{6mm} $featureVector \leftarrow concatValues(featureVector, binaryVal)$ \>\\

\hspace{9mm} \tiny{// Add feature vector to feature matrix} \\
13:~ \hspace{3mm} $\mathbf{F}_x \leftarrow concatToMatrix(triple, featureVector, \mathbf{F}_x)$ \>\\
 
14:~ \textbf{return} $\mathbf{F}_x$
\\[-0.8cm]
\end{tabbing}
\end{algorithm*}

For entity anomaly detection, in addition to structural features, we also generate content-based binary features 
by considering data quality aspects of the literals such as the 
presence/absence, validity/invalidity, and so on, of a triple. This way, we can identify abnormal entities in a KG considering both 
structure and content. Algorithm~\ref{alg:structuralfeatures} provides the pseudocode of how we generate structural features, while Algorithm~\ref{alg:contentfeatures} provides the pseudocode of the process of generating content-based features.

\begin{algorithm*}[t!]
\scriptsize
\caption{Generation of content-based features}
\label{alg:contentfeatures}
\SetAlgoNoLine
\KwIn{$G$: \hspace{1.4mm} The knowledge graph.}
\KwOut{$dictFeatureVector$: \hspace{0.1mm} The dictionary of content-based feature vectors.}
\smallskip
\begin{tabbing}
0:~ \hspace{3mm} $\mathbf{C} \leftarrow
    setfeaturesfornodesfromattributetypeandkindofanyof(G, F, L)$~\= \kill
\hspace{3mm} \tiny{// Fetch triples from $G$, where $o$ is a literal} \\
1:~ $G_l \leftarrow getTriples(G)$ \> \\        

\hspace{3mm} \tiny{// Dictionary with each entity's literal count, where $key=entity$ and $value=countLiterals$} \\
2:~ $dictLiteralCounts \leftarrow getLiteralCountEachNode(G_l)$ \>  \\      

\hspace{3mm} \tiny{// Get the median value of the literal counts} \\
3:~ $medianLiteralCount \leftarrow getMedOfCounts(dictLiteralCounts.values)$ \> \\         

\hspace{3mm} \tiny{// Iterate over each triple in $G_l$} \\
4:~ \textbf{for} $triple$ in $G_l$: \> \\

\hspace{10mm} \tiny{// Get the $s$, $p$ and $o$ of a triple} \\
5:~ \hspace{5mm} $s,p, o \leftarrow getSubject(triple), getPredicate(triple), getObject(triple)$ \> \\

\hspace{10mm} \tiny{// Check if literal value is present} \\
6:~ \hspace{6mm}$valuePresence \leftarrow checkEmpty(o)$\> \\

\hspace{10mm} \tiny{// Check if literal matches predicate meaning} \\
7:~ \hspace{6mm}$validLiteral \leftarrow literalMatchesPredicate(p, o)$\> \\

 \hspace{10mm} \tiny{// Check if $s$, $p$ together occur more than once} \\
8:~ \hspace{6mm}$redundantInfo \leftarrow CountOccurence(s, p)$\>\\

\hspace{10mm} \tiny{// Get count of triples associated with $s$}\\
9:~ \hspace{6mm}$literalCount \leftarrow getCountOfLiterals(s, G_l)$\>\\

\hspace{10mm} \tiny{// Check if entity has high count of facts} \\
10:~ \hspace{5mm}$richEntity \leftarrow freqCountLiterals(literalCount, medianLiteralCount)$\> \\

\hspace{10mm}  \tiny{// Construct feature vector with binary features}\\
11:~\hspace{6mm}$ featureVect \leftarrow constFVector(valuePresence, validLiteral$,$redundantInfo, richEntity) $\> \\

\hspace{6mm}\tiny{// Add feature vector to dictionary, where $key=s$ and $value=featureVect$}\\
12:~\hspace{1mm}$ dictFeatureVector \leftarrow addFeatureVectors(s, featureVect) $\> \\

\hspace{6mm}\tiny{//This output is used in constructing $\mathbf{F}_e$} \\
13:~ \textbf{return} $dictFeatureVector$ \>\\
    
\\[-0.8cm]
\end{tabbing}
\end{algorithm*}

We next train a one-class Support Vector Machine (SVM)~\cite{scholkopf2001svm}
to perform unsupervised anomaly detection on the generated feature
matrices.

\subsection{Learning a One-Class SVM}

A one-class
$\nu$-SVM~\cite{scholkopf2001svm} learns a non-linear decision function where data points that
have a non-negative distance, $0 \le d$, from the decision boundary, are
in the region capturing the majority of data points. This is considered
as the \emph{normal} class. Negative distances, $d < 0$, identify data
points outside this region and are considered to be the \emph{abnormal}
class. A one-class $\nu$-SVM achieves this separation using a kernel
function~\cite{christen2016anomalydetection, vapnik2000statisticallearning}. 

To identify the most abnormal
feature vectors, we train several one-class $\nu$-SVMs using
different kernel functions (such as linear function, RBF, polynomial function, and sigmoid function). For each $\nu$-SVM we train, we set $\nu=
b / n$, where $n = |\mathbf{F}|$. This is the fraction of vectors in
the feature matrix $\mathbf{F}$, that a domain expert will be able to
investigate, where $1 < b \le |\mathbf{F}|$, and it is named as the
\emph{budget}. As we input the feature matrix $\mathbf{F}$ to
one-class $\nu$-SVM that is generated in step (\textbf{b}), the
$\nu$-SVM returns a numerical distance $\mathbf{f}.d$ from the
decision boundary for each feature vector $\mathbf{f} \in
\mathbf{F}$~\cite{senaratne2022anomalnodes}.

We then classify feature vectors as normal if their distance is
$\mathbf{f}.d \ge 0$ or abnormal if their distance is $\mathbf{f}.d <
0$. Specifically, we generate the set of normal feature vectors,
$\mathbf{N} = \{\mathbf{f} \in \mathbf{F}: 0 \le \mathbf{f}.d\}$ and
abnormal feature vectors, $\mathbf{A} = \{\mathbf{f} \in \mathbf{F}: 0
> \mathbf{f}.d\}$, where $\mathbf{F} = \mathbf{N} \cup \mathbf{A}$. If we train the $\nu$-SVM with $\mathbf{F}_x$, then the output will be the
set of anomalous facts, and with $\mathbf{F}_y$, the output will be the
set of anomalous entities~\cite{christen2016anomalydetection}. 

Once we have
identified anomalies, we can remove the identified
abnormal triples from the KG and use the refined graph in a 
downstream task such as Knowledge Graph Completion (KGC), or we can perform manual validation with the involvement of domain experts.\newline

\subsection{Entity Type Generation for Anomaly Detection}
\label{subsec:entgene}

While most KGs provide type information for entities, many KGs still lack this type of information. Therefore, we introduce
ENTGENE (\underline{En}tity \underline{T}ype \underline{Gene}rator), which operates independently from other
external KGs to generate or infer the entity types in their absence. For entity type determination, ENTGENE uses Named
Entity Recognition (NER) in NLP~\cite{nadeau2007ner}, and ENTGENE helps to detect any abnormal use of predicates with a subject or entity type.

For example, using the predicate \emph{livesIn} with the object
\emph{1993-10-10} is incorrect as the object should be of the
type \emph{LOCATION}, not \emph{DATE} type. If the type information is partially available in the KG, or when
the type information cannot be generated (identified as \emph{OTHER} by NER), ENTGENE infers the type using the type information of already identified entities. 

For example, if the majority of the objects $o \in \mathbf{I}_e$
in triples with the predicate \emph{livesIn} have the type
\emph{LOCATION}, ENTGENE makes this inference for entities with missing type
information, where the object of a triple with the predicate
\emph{livesIn} should be of type \emph{LOCATION}. While it is possible to
have multiple types associated with an entity due to inferencing, any
abnormalities related to entity type information will be identified
during the process of entity anomaly detection by SEKA.

\section{TAXO: Taxonomy of Knowledge Graph Anomalies}
\label{sec:taxo}

TAXO considers possible anomaly types that can occur either in an RDF storage file such as Terse RDF Triple Language (Turtle)\footnote{\label{fn3}\url{https://www.w3.org/TR/turtle/}}, and anomaly types that we can discover upon graph population.
We primarily classify anomaly types based on the type of a triple, where we separate the anomalies occurring in triples of the form $\mathbf{I}_e \times \mathbf{I}_r \times \mathbf{I}_e$ (entity-to-entity) from the anomalies occurring in triples of the form $\mathbf{I}_e \times \mathbf{I}_r \times \mathbf{L}$ (entity-to-literal). This 
separation is useful as the anomaly detected and the method of correction 
differs based on the triple type, and not every triple type has the same type of anomalies. For each triple type, we further classify anomalies based on the number of triples involved in the anomaly 
as single or multiple. The solutions to rectify the identified anomalies can be either one of the following three types.
\begin{enumerate}
    \item Automatic correction: This refers to the use of tasks such as link prediction~\cite{lu2019heterolinkpred} or entity disambiguation~\cite{cucerzan2007entitydisambiguation}, where we can use an automated technique performing these tasks to carry out the correction of an anomaly.
    \item Human evaluation: This refers to obtaining human involvement for KG refinement. If there is no automatic means of correcting an anomaly, or if the anomaly is worthwhile receiving human inspection, we can forward triples for evaluation by domain experts.
    \item Remove RDF entry: This method of elimination suggests removing the triple from the graph, when there is no automatic method to correct them, such a technique is not required, when it is not worth requesting human intervention, or the triple is a non-erroneous outlier that adds no knowledge such as duplicates or redundant triples.
\end{enumerate}

While the above three types of corrections may heal or modify a KG, what is important to consider is the impact the proposed technique makes on the coverage of the KG. Even though RDF removal seems to reduce the coverage of a KG, with our experimental results provided in Section~\ref{subsec:experimentsseka} where we demonstrated how the removal of abnormal triples (over random removal) can positively influence downstream tasks such as link prediction. Hence, triple removal does not always make a negative impact towards KG coverage.

\begin{table*}[t!]
\centering
\caption[Anomaly types involving entity-based literals.]{Anomaly types involving entity-based literals with examples extracted from YAGO-1, DSKG, Wikidata, and KBpedia.}
\label{tab:taxoentity}
\scalebox{0.57}{
\begin{tabular}{llllll}
    \toprule
    \shortstack[l]{Triples \\involved} & \shortstack[l]{Anomaly\\type} & \shortstack[l]{Anomaly\\name} & \multirow{-2}{*}{Example} & \shortstack[l]{Anomaly\\source} & Possible correction \\
    \midrule
    \multirow{8}{*}{\shortstack[l]{Single\\triple}} & \multirow{3}{*}{Missingness} & \shortstack[l]{Missing\\subject} & \multirow{-2}{*}{( , bornIn, London)}  & \shortstack[l]{Storage\\file} & \multirow{-2}{*}{Link prediction.}\\
    \cline{3-6}
    \\[-2.2ex]
    & & \shortstack[l]{Missing\\predicate} & \multirow{-2}{*}{(9thWonder,  , TheDreamMerchantVol2)} &  \shortstack[l]{Storage\\file} & \multirow{-2}{*}{Link prediction.} \\
    \cline{3-6}
    \\[-2.2ex]
    & & \shortstack[l]{Missing\\object} & \multirow{-2}{*}{(Distribution/100150, byteSize, )} &  \shortstack[l]{Storage\\file} & \multirow{-2}{*}{Link prediction.}\\
    \cline{2-6}
    \\[-2.2ex]
    & Incorrectness & \shortstack[l]{Incorrect\\predicate} & \multirow{-2}{*}{(MarlaMaples, hasChild, DonaldTrump)} &  \multirow{-2}{*}{Graph} & \multirow{-2}{*}{Link prediction.}\\
    \cline{2-6}
    \\[-2.2ex]
    & Inconsistency & \shortstack[l]{Invalid\\predicate} & \multirow{-2}{*}{(Stavisky, isLocatedIn, France)} &  \multirow{-2}{*}{Graph} & \multirow{-2}{*}{Link prediction.}\\
   \cline{2-6}
   \\[-2.2ex]
    &\multirow{6}{*}{Ambiguity} & \multirow{3}{*}{\shortstack[l]{Entity\\ambiguity}} & (JoshGracin, originatesFrom, JoshGracin) & Graph & Entity disambiguation \\
     \cline{1-1}
     \cline{4-6}
     \\[-2.2ex]
    \multirow{14}{*}{\shortstack[l]{Multiple\\triples}} &  &   & (Marcelona, bornIn, Mozambique) &  \multirow{2}{*}{Graph} & \multirow{2}{*}{Entity disambiguation.}\\
    & & & (Marcelona, hasSuccessor, Mozambique) \\
    \cline{3-6}
    \\[-2.2ex]
    & & \multirow{3}{*}{\shortstack[l]{Predicate\\ambiguity}} &(Ain'tTooProudToBeg, isOfGenre, rock)  & \multirow{3}{*}{Graph} & \multirow{3}{*}{\shortstack[l]{Human evaluation, or\\KG re-engineering.}}\\
    & & & (Ain'tTooProudToBeg, isOfGenre, music) \\
    & & & (Ain'tTooProudToBeg, isOfGenre, popularMusic) \\
    \cline{2-6}
    \\[-2.2ex]
    & \multirow{2}{*}{Contradictions} & \multirow{2}{*}{\shortstack[l]{Contradicting\\facts}} & (DonaldTrump, marriedTo, MarlaMaples) &  \multirow{2}{*}{Graph} & \multirow{2}{*}{\shortstack[l]{Remove incorrect\\RDF entry.}}\\
    & & & (MarlaMaples, hasChild, DonaldTrump) \\
    \cline{2-6}
    \\[-2.2ex]
    & \multirow{2}{*}{Unusual} & Rare entity & Entity \textit{9thWonder} has only one fact &  Graph & No correction required. \\
    \cline{3-6}
    \\[-2.2ex]
    & &  Prolific entity & Entity \textit{Dataset/410} has 11 \textit{createdBy} links &  Graph & No correction required. \\
    \cline{2-6}
    \\[-2.2ex]
    & \multirow{2}{*}{Redundancies} & \multirow{2}{*}{\shortstack[l]{Redundant\\facts}} & (AMGRapper, produced, BettaHaveMoney) &  \multirow{2}{*}{Graph} & \multirow{2}{*}{Human evaluation.}\\
    & & & (AMGRapper, produced, BettaHaveMoney2001) \\
    \cline{2-6}
    \\[-2.2ex]
    & \multirow{2}{*}{Duplicates} & \multirow{2}{*}{\shortstack[l]{Duplicate\\facts}} & (LizBeth, bornIn, Mexico)  &  \multirow{2}{*}{\shortstack[l]{Storage\\file}}  & \multirow{2}{*}{Remove duplicates.}\\
    & & & (LizBeth, bornIn, Mexico)\\    
    \bottomrule
\end{tabular}}
\end{table*}

In Table~\ref{tab:taxoentity}, we identify eight different types of anomalies that can occur involving 
a triple of the form $\mathbf{I}_e \times \mathbf{I}_r \times \mathbf{I}_e$ (entity-to-entity), 
where both its subject and object are entities. In Table~\ref{tab:taxoliteral}, we identify five types of anomalies that can occur involving one or multiple 
triples of the form entity-to-literal, where 
the subject of such a triple is an entity, and the object is a literal. A literal can be a string, date, number, or hyperlink. Anomalies such as a 
missing element, incorrectness, and type inconsistencies, usually 
involve a single triple. Whereas, anomalies such as ambiguity, 
contradictions, redundancies, and duplicates involve multiple 
triples. Interestingness can involve triples or entities that are 
classified as abnormal, but non-erroneous.

\begin{table*}[t!]
\centering
\caption[Anomaly types involving literal-based triples.]{Anomaly types involving literal-based triples with examples extracted from YAGO-1, DSKG, Wikidata, and KBpedia.}
\label{tab:taxoliteral}
\scalebox{0.61}{
\begin{tabular}{lllllll}
    \toprule
    \shortstack[l]{Triples \\involved} & \shortstack[l]{Anomaly\\type} & \shortstack[l]{Anomaly\\name} & \multirow{-2}{*}{Example} & \shortstack[l]{Anomaly\\source} & \multirow{-2}{*}{Possible correction} \\
    \midrule
    \multirow{7}{*}{\shortstack[l]{Single\\triple}} & \multirow{3}{*}{Missingness} & \shortstack[l]{Missing\\subject} & \multirow{-2}{*}{( , hasDefinition, "A query language")} &  \shortstack[l]{Storage\\file} & \multirow{-2}{*}{Remove RDF entry.}\\
    \cline{3-6}
    \\[-2.2ex]
    &  & \shortstack[l]{Missing\\predicate} & \multirow{-2}{*}{(Echidna,  , "Egg laying mammal")} &  \shortstack[l]{Storage\\file} & \multirow{-2}{*}{Link prediction.} \\
    \cline{3-6}
    \\[-2.2ex]
    & & \shortstack[l]{Missing\\literal} & \multirow{-2}{*}{(SQL, hasDefinition, "")} &  \multirow{-2}{*}{Graph} & \multirow{-2}{*}{Link prediction.}\\
    \cline{2-6}
    \\[-2.2ex]
    & \multirow{3}{*}{Incorrectness} & \multirow{2}{*}{\shortstack[l]{Incorrect\\literal}} & (Aristotle, bornOn, "380") &  \multirow{2}{*}{Graph} & \multirow{2}{*}{Human evaluation.}\\
    & &  & (DJShadow, created, "album") &  &   \\
    \cline{3-6}
    \\[-2.2ex]
    &  & \shortstack[l]{Partially\\correct\\literal} & \multirow{-2}{*}{(AliHewson, bornOn, "1961-\#\#-\#\#")} &  \multirow{-2}{*}{Graph} & \multirow{-2}{*}{Human evaluation.}\\
    \cline{2-6}
    \\[-2.2ex]
    & Inconsistency & \shortstack[l]{Invalid\\predicate} & \multirow{-2}{*}{(AMGAlbum, bornOn, "2001-11-25")} &  \multirow{-2}{*}{Graph} & \multirow{-2}{*}{Human evaluation.} \\
    
    \cline{1-6}    
    \\[-2.2ex]
    \multirow{7}{*}{\shortstack[l]{Multiple\\triples}} &  \multirow{5}{*}{Redundancies} & \multirow{5}{*}{\shortstack[l]{Redundant\\literals}}  & (SQL, isa, "Programming Language") &  \multirow{5}{*}{Graph} & \multirow{5}{*}{Human evaluation.}\\
    & & & (SQL, isa, "Programming-Language") \\
    \cline{4-4}
    \\[-2.2ex]
    & & & (Zucchini, altLabel, "fruit of zucchini plant") \\
    & & & (Zucchini, altLabel, "fruit of zucchini plants") \\
    & & & (Zucchini, altLabel, "fruit of zucchini") \\
    \cline{2-6}
    \\[-2.2ex]
    & \multirow{2}{*}{Duplicates} & \multirow{2}{*}{\shortstack[l]{Duplicate\\facts}} & (LizBeth, bornOn, "1948-04-20") &  \multirow{2}{*}{Graph} & \multirow{2}{*}{Remove duplicates.}\\
    & & & (LizBeth, bornOn, "1948-04-20") \\
    \bottomrule
\end{tabular}}
\end{table*}

\subsection{Missingness}
\label{subsec:missingness}

Given that a triple is composed of three elements (subject, predicate, object), one of the basic anomalies that can occur is the absence of either one of these elements. That is either the subject (head), predicate (relationship), or object (tail) can be missing. This can happen due to an anomaly in the source from which the data was extracted, an anomaly in the triples extractor, or human failures. Even though such triples with missing elements can occur in the KG storage file formats such as Turtle\footref{fn3}, we can use link prediction~\cite{lu2019heterolinkpred} to resolve such missing elements.

Furthermore, with validation techniques such as SHACL~\cite{delva2023shacl}, the identification of such missing elements within a triple is not a difficult task. Even if a KG does not have such a validation layer or phase, triples with missing elements will be ignored at the time of the KG population. Thus, such triples do not affect the performance of downstream tasks such as link prediction. For example, the triple \textit{( , bornIn, London)} given in Table~\ref{tab:taxoentity} is an example of missingness that we extracted from the graph storage files.

Similar to entity-to-entity triples, entity-to-literal triples can also have a missing element in the triple recorded in the storage file. For example, consider the triple \textit{( , hasDefinition, "A query language")} in Table~\ref{tab:taxoliteral}. However, in contrast to a missing object of an entity-to-entity triple, a missing object of an entity-to-literal occurs at the graph level too. Even though these triples with a missing element can be detected by a validation technique, it is important that they are corrected instead of a mere triple deletion as this affects the coverage of a KG. Hence, scenarios such as missing predicates and literals can be corrected with link prediction~\cite{lu2019heterolinkpred}. Furthermore, we can also get expert evaluation to detect missing predicates.

\subsection{Incorrectness}

In TAXO, an incorrect triple is one that states a wrong relationship between two entities. In an unsupervised setting that is independent of external resources, such incorrect triples can only be identified if another triple becomes anomalous due to the occurrence of this incorrect triple. For example, consider the incorrect triple \textit{(MarlaMaples, hasChild, DonaldTrump)} given in Table~\ref{tab:taxoentity}. This incorrect triple was discovered as a result of the contradiction it makes with the triple \textit{(DonaldTrump, marriedTo, MarlaMaples)}. While removing this incorrect triple will remove the incorrect RDF entry, correction of an incorrect triple requires human intervention or inputs from an external source. If the two corresponding entities have no such relationship, then the solution can be as simple as edge removal. However, if two entities share a relationship in the real-world, link prediction can be the solution when human intervention is not a possibility.

For literal-based triples, we can decide if the literal value is completely or partially correct with the support of a domain expert, or via comparison with an external source.
An incorrect triple in this scenario is one with an incorrect literal. As per the example given in Table~\ref{tab:taxoliteral}, the birth date of \textit{Aristotle} is given as \textit{380}. While he was born in 384 BC, a mere number in place of a date is considered wrong as the literal value does not match with the data types implied by the predicate of the triple. Identifying incorrect entity-to-literal triples is easier than detecting incorrect entity-to-entity triples. However, detecting incorrect triples where the literals do not have an error that is easily identifiable requires expert evaluation or comparison with an external source.

Partially correct triples are an anomaly type that can be seen only in entity-to-literal triples.  For example consider the triple \textit{(JohnHanbury, diedOnDate, "1734-\#\#-\#\#")}. While it is true that \textit{John Hanbury} died in 1734, the triple is classified partially correct as it does not contain information about the exact date he passed away. Similar to completely incorrect triples, the detection of partially correct triples requires expert inputs, or inputs from external sources.

\subsection{Inconsistency}

Similar to incorrect triples, an inconsistent triple is one 
that also becomes wrong because of an invalid predicate used with a particular subject and object. For example, consider the inconsistent triple \textit{(Stavisky, isLocatedIn, France)} given in Table~\ref{tab:taxoentity}. While \textit{France} is a \textit{LOCATION}, the entity \textit{Stavisky} refers to the name of a drama and a financial scandal that took place in France in 1934. While the predicate \textit{isLocatedIn} is mostly used with a subject of the type \textit{ORGANIZATION} or \textit{LOCATION}, together with an object of the type \textit{LOCATION}, the abnormality with this triple is the unusual use of the predicate with an entity that is neither an organization, nor a location.

Furthermore, an incorrect usage of a predicate will create an invalid triple. Consider the triple \textit{(AMGAlbum, bornOn, "2001-11-25")} given in Table~\ref{tab:taxoliteral}. While \textit{AMG} is an album, it should have a date of release or creation instead of a birth date. Similar to the type inconsistency anomaly occurring in entity-to-entity triples, in entity-to-literal triples also we need to assess the entity type of the subject, usage of the predicate, and the expected data type of the literal by the predicate to determine the validity or invalidity of the predicate. While edge removal is a possibility, any possible corrections to the literal will require expert input.

It is possible to argue that an invalid triple can also be classified as an incorrect one. We classify a triple as invalid on the basis that the predicate has an invalid usage. For example, a predicate that should be used with people being used with a location is considered invalid. In the scenario of literal-based triples, as per the triple \textit{(AMGAlbum, bornOn, "2001-11-25")} in Table~\ref{tab:taxoliteral}, while it is considered invalid for an album to have a date of birth, the triple is also considered incorrect if the given date of creation of the album is incorrect. Hence, an invalid triple has the potential of becoming incorrect too.

\subsection{Ambiguity}

Ambiguous triples can either have an ambiguity with the entity, 
or with the usage of a predicate. For example, consider the 
ambiguous triple \textit{(JoshGracin, originatesFrom, JoshGracin)} given in Table~\ref{tab:taxoentity}. In this triple, the subject refers to a 
music album while the object refers to a person. \textit{Josh Gracin} is the name of the debut album released by \textit{Josh Gracin} in 2004. This entity ambiguity also leads to inconsistency as a person and a music album are mostly related by predicates such as \textit{produced/producedBy}, or \textit{created/createdBy}. Similarly, consider the two triples \textit{(Marcelona, bornIn, Mozambique)} and \textit{(Marcelona, hasSuccessor, Mozambique)}. While \textit{Mozambique} in the first triple refers to a country, in the second triple it refers to a person. As the same entity name has been used to refer to entities of two different types, entity disambiguation~\cite{cucerzan2007entitydisambiguation} 
is required to solve such scenarios. 

Furthermore, ambiguity can occur in a predicate where the same predicate 
is being used with the same subject but with different objects, thus 
making the range of the predicate unclear. Consider the three triples \textit{(Ain'tTooProud ToBe, isOfGenre, rock)}, \textit{(Ain'tTooProudToBe, isOfGenre, music)}, and \textit{(Ain't TooProudToBe, isOfGenre, popularMusic)} given in 
Table~\ref{tab:taxoentity}. As this ambiguity in the predicate usage may not 
necessarily be erroneous, correction will require human 
intervention as some of these triples may also cause redundancies. In the worst-case scenario, KG re-engineering~\cite{haakansson2006kgreengineering} will be required to specify the range of predicates and establish classes among entities to denote any inheritance among them.

\subsection{Contradictions}

TAXO considers a contradiction among triples to happen when two 
entities share two different predicates that cannot co-exist in the real-world. As per the two triples \textit{(DonaldTrump, marriedTo, MarlaMaples)} and \textit{(MarlaMaples, hasChild, Donald Trump)} given in 
Table~\ref{tab:taxoentity}, one person being both the child and 
spouse of another person is considered erroneous as it is 
highly unlikely to happen in the current world. In the 
scenarios of a contradiction, out of the triples involved, there is a 
possibility of some or all of them becoming wrong triples. Removal of the 
incorrect triple(s) will remove the contradiction among the triples. 

However, it can be argued that a contradiction may also lead to an entity 
ambiguity. Considering the example in Table~\ref{tab:taxoentity}, the 
contradiction is when we consider \textit{Donald Trump} and \textit{Marla Maples} in the 
first triple to be the same people in the second triple. But, if we 
consider these individuals to be four different or three different people, then 
these two triples reflect the scenario of ambiguous entities via a contradiction.

Similar to anomalies that involve either entity-based (such as the examples in Table~\ref{tab:taxoentity}) or literal-based triple(s) (such as the examples in Table~\ref{tab:taxoliteral}), there are anomalies that occur due to contradictions between triples from both these types. Hence, we name this anomaly as mixed triples. While multiple triples together form this anomaly, it is possible that all or one of the involved triples convey incorrect information, thus creating a contradiction. For example consider the two triples given in Table~\ref{tab:taxoentityliteral}. While it is true that \textit{Donald Trump} has a son by the name \textit{Donald Trump Junior}, the contradiction occurs when associating the son's name as an alternative name for the father, \textit{Donald Trump}.

\subsection{Unusual}

What is to be 
called unusual or interesting is application and domain-dependent. However, 
we consider entities with characteristics that are 
different from the rest of the triples under consideration as 
being unusual. Unusual entities need not be erroneous at all times, and thus do not require a correction. But, they have 
the potential to act as a source of knowledge. As per 
Table~\ref{tab:taxoentity}, we consider two types of interesting 
entities. We name entities with only one or a few facts as rare entities, and 
entities with a count of triples (or triples of a specific 
type) that is comparatively higher than the other entities of 
the same type as prolific entities. 

With reference to the two examples in 
Table~\ref{tab:taxoentity}, the entity \textit{9thWonder} is 
considered rare as it is an entity with just one fact. The 
entity \textit{Dataset/410} is considered prolific as this 
dataset has eleven triples with the predicate 
\textit{createdBy}, whereas other datasets in DSKG have at most 
five creators. Even though this scenario does not seem to
be erroneous, it is still possible that \textit{Dataset/410} 
has creators of other datasets mistakenly related to it.

\subsection{Redundant Triples}

A redundancy is when two or more triples seem to convey the 
same information, in which sometimes the information provided 
can also be contradicting. With reference to the example in 
Table~\ref{tab:taxoentity}, \textit{BettaHaveMoney} is an album 
produced by \textit{AMGRapper}. While both the triples in the 
provided example convey this information, what requires human 
intervention is to determine whether \textit{BettaHaveMoney} 
and \textit{BettaHaveMoney2021} refer to the same or different \
albums. If they are the same, then one of the triples can be 
safely removed after resolving the respective entity in the 
object. Hence, we suggest forwarding such scenarios to domain 
experts for evaluation.

While redundancies in entity-to-entity triples involve redundant entities, in entity-to-literal a redundancy involves a literal value. Given a literal value can be a string as short as one character, or as long as 100 characters, it is easier to have a high number of redundancies with literal-based triples (due to a high number of character variations we can have in a literal) in comparison to the number of redundancies involved with entity-based triples. Refer to the examples of \textit{SQL} and \textit{Zucchini} given in Table~\ref{tab:taxoliteral}. All these triples convey the same information. 

The reason for having multiple triples is the slight differences in the strings provided as the object. A slight difference in punctuation or spelling is the main reason for having redundant literals. In addition, we have also encountered literals where the date of birth of people are mentioned in both date format (09/11/1964) and words (09th of November 1964), where such scenarios also lead to the provision of redundant information. While these anomalies occur in a KG, removing them will improve the quality of the KG, whilst positively contributing towards the performance of KG validation tools.

\begin{table*}[t!]
\centering
\caption[Anomaly types involving both entity-based and literal-based triples.]{Anomaly types involving both entity-based and literal-based triples with an example extracted from YAGO-1.}
\label{tab:taxoentityliteral}
\scalebox{0.62}{
\begin{tabular}{llllll}
    \toprule
    \shortstack[l]{Triples \\involved} & \shortstack[l]{Anomaly\\type} & \shortstack[l]{Anomaly\\name} & Example & \shortstack[l]{Anomaly\\source} & Possible correction \\
    \midrule
    \multirow{2}{*}{\shortstack[l]{Multiple\\triples}} & \multirow{2}{*}{Contradiction} & \multirow{2}{*}{\shortstack[l]{Mixed\\triples}} & (DonaldTrump, hasChild, DonaldTrumpJr.) & \multirow{2}{*}{Graph} & \multirow{2}{*}{Human evaluation.}\\
    & & & (DonaldTrump, altLabel, "Donald Trump Jr.,")\\
    \bottomrule
\end{tabular}}
\end{table*}

\subsection{Duplicates}

While there will not be any duplicates in entity-based triples, duplicates may still occur in the graph storage files. While such duplication can create disadvantages to decision-makers and as well as to processing techniques, removing the duplicates will enhance the quality of the graph storage files. Thereby, providing a cleansed input to the KG validation and enrichment tools, and improving their performance. At present, a real-world KG with a validation layer formed by ShEx or SHACL~\cite{rabbani2023shexshacl} will have such duplicates easily identified. However, it is important to note that not every KG may have these validation techniques adopted.

Even though duplicate entity-based triples exist only in the KG storage file, duplicate literal-based triples can occur in the KG as well. While the solution of resolving these duplicates can be as simple as removing the duplicates leaving only a single entry, it will be beneficial to take necessary precautions at the time of KG construction to prevent any such duplicates from being added as they do not add any new knowledge. Furthermore, the existence of duplicates that add no value makes graph storage files large, thus adding a burden during KG transfer and processing.

\pagebreak

\section{Computational Complexity}
\label{sec:complexity}

In SEKA, the step of generating features which is dependent on the CPA
algorithm, uses depth-first search to generate the paths. A
single path ($k=1$) can be found in $O(V+E)$ time, where $V$
corresponds to the nodes and $E$ corresponds to the edges. However,
the number of simple paths in a graph can be as large as $O(m!)$ in
the entire graph of order $m$. The complexity of 
training a $\nu$-SVM is quadratic ($O(E^2)$) in the number of training records used.

\section{Experimental Evaluation}
\label{sec:experimentaleval}

We use the four real-world KGs YAGO-1\footnote{\label{fn4}\url{https://yago-knowledge.org/downloads/yago-1}},
KBpedia\footnote{\label{fn5}\url{https://kbpedia.org/}},
Wikidata\footnote{\label{fn6}\url{https://www.wikidata.org/wiki/Wikidata:Main\_Page}}, and DSKG\footnote{\label{fn7}\url{http://dskg.org/}} to evaluate the path-based approach, and to conduct experiments with the baselines. We selected the KGs in such a way that they are of different sizes and belong to different domains, and are of different qualities. As the four real-world KGs do not have labelled data, we generated synthetic corruptions in the KGs using TRIC~\cite{senaratne2023tric}. The source code of SEKA is available on GitHub\footnote{\url{https://github.com/AsaraSenaratne/SEKA}}. 

\subsection{Parameter Settings}
\label{sec:parametersettings}
We now discuss the parameter settings used in SEKA. To eliminate the need
of having user involvement during the anomaly detection process, our
approach is designed in such a way that it has a minimal number of
parameters that need to be manually set.

In constructing features, to determine the path length to traverse in depth-first search, we use $k$ where it is by default set to $k = 2$. Hence,
no user involvement is required to determine the value of $k$.

The two most important parameters to configure in training the $\nu$-SVM are its kernel and $\nu$. In order to obtain robust results with 
regard to which feature vectors in $\mathbf{F}$ are normal and abnormal, we train several $\nu$-SVMs with different kernels. This ensures our classification outcomes
are not biased due to the use of a single kernel function. The ratio
of normal to abnormal classified feature vectors can be specified
using the $\nu$ parameter, which can be set by a user according to
the budget $b$ of how many abnormal feature vectors she/he can
investigate further. By default, $b$ is set to 100 assuming this is the number of triples an expert can handle manually.

Next, we provide an extensive evaluation of SEKA's performance. First, we experiment using synthetically generated anomalies, where we compare SEKA with its baselines. Next, we perform anomaly detection on the original KGs, where we evaluate the results manually. Finally, we apply anomaly detection in a downstream task such as KGC to show how anomaly detection can complement such downstream tasks.

\subsection{Anomaly Detection With Synthetic Anomalies}
\label{subsec:experimentsseka}

As the four real-world KGs do not have labelled data, we use TRIC~\cite{senaratne2023tric} to synthetically
generate anomalies for each KG by manually corrupting the triples. We compare SEKA with the following three baseline approaches. We use precision and recall values to determine how well each approach performs in identifying the anomalies. 

\begin{enumerate}
    \item PaTyBRED (Paths and Types with Binary Relevance for Error
    Detection)~\cite{melo2017detectrelationassertions} is a method used
    for the detection of relation assertion errors in KGs, which incorporate type and path features into local
    relation classifiers. 
    \item SDValidate~\cite{paulheim2014improvingqualityofkg} relies on statistical distributions of types and relations, such
    as characteristic distributions of the types of a property's
    subject and object, and applies outlier detection to detect
    erroneous relation assertions.
    \item KGTtm (Knowledge Graph Triple trustworthiness
    measurement)~\cite{jia2019trustworthinesskg} synthesizes the internal semantic
    information in the triples and the global inference information
    of the KG to achieve the trustworthiness measurement and fusion
    in the three levels of entity level, relationship level, and KG
    global level. 

\end{enumerate}

As shown in Table~\ref{tab:baselinecomp},  which presents the experimental results with 10\% of the triples corrupted, SEKA outperforms all the baselines under consideration. Compared to these baselines, SEKA achieves an increase of up to 0.10, 0.12, 0.11, and 0.12 in precision, and an increase of up to 0.12, 0.13, 0.12, and 0.11 in recall for YAGO-1, DSKG, Wikidata, and KBpedia, respectively. Thus, SEKA demonstrates superior performance in identifying anomalous triples. Similarly, SEKA outperforms the three baselines when 20\% and 30\% of the triples are corrupted.

\begin{table*}[t!]
\caption[Comparative evaluation of SEKA with the baselines.]{Comparison of SEKA with baselines with 10\% of the triples corrupted. The best results are shown in bold.}
\begin{center}
\scalebox{0.75}{
\begin{tabular}{lrrrrrrrr}
    \toprule
    \multirow{2}{*}{Approach}& \multicolumn{2}{c}{YAGO-1}  & \multicolumn{2}{c}{DSKG} & \multicolumn{2}{c}{Wikidata} & \multicolumn{2}{c}{KBpedia}\\
    \cline{2-9}
    & Precision & Recall & Precision & Recall & Precision & Recall & Precision & Recall\\
    \midrule
    PaTyBRED & 0.87 & 0.86 & 0.84 & 0.83 & 0.72 & 0.72 & 0.77 & 0.75 \\ 
    SDValidate & 0.82 & 0.80 & 0.81 & 0.80 & 0.70 & 0.68 & 0.71 & 0.71\\
    KGTtm & 0.86 & 0.84	& 0.88 & 0.83 & 0.77 & 0.77 & 0.70 & 0.69 \\
    \textbf{SEKA} & \textbf{0.92} & \textbf{0.92} & \textbf{0.93} & \textbf{0.93} & \textbf{0.81} & \textbf{0.80} & \textbf{0.82} & \textbf{0.81}\\
    \bottomrule
\end{tabular}}
\end{center}
\label{tab:baselinecomp}
\end{table*}

\begin{table}[t!]
\caption[Comparison of performance between PRA and CPA.]{Comparison of performance between PRA and CPA for general fact anomaly detection with 10\% of triples corrupted. The best results are shown in bold.}
\begin{center}
\scalebox{0.9}{
\begin{tabular}{llrrr}
    \toprule
    {KG} & Approach & Run time (min) & Precision & Recall\\
    \midrule
    \multirow{2}{*}{YAGO-1} & PRA & 178 & 0.85 &  0.84\\
     & \textbf{CPA} & \textbf{121} & \textbf{0.92} & \textbf{0.92}\\
    \hline
    
    \multirow{2}{*}{KBpedia} & PRA & 88 & 0.80 & 0.79 \\
     & \textbf{CPA} & \textbf{48} & \textbf{0.82} & \textbf{0.81} \\
    \hline
    
    \multirow{2}{*}{Wikidata} & PRA & 301 & 0.79 & 0.78 \\
     & \textbf{CPA} & \textbf{232} & \textbf{0.81} & \textbf{0.80}\\
    \hline
    
    \multirow{2}{*}{DSKG} & PRA & 137 & 0.92 &  0.90\\
    & \textbf{CPA} & \textbf{72} & \textbf{0.93} & \textbf{0.93}\\
    \bottomrule
\end{tabular}}
\end{center}
\label{tab:CPAnpra}
\end{table}

To evaluate CPA versus PRA, we conduct
experiments to assess the run time, and the quality of the anomalies detected by
probabilistic feature generation versus binary feature generation.
In Table~\ref{tab:CPAnpra}, we show the experimental results
obtained with 10\% of the triples corrupted in the four KGs. As can be seen from Table~\ref{tab:CPAnpra}, CPA outperforms PRA on all four KGs with substantially reduced run times and higher precision and recall values. This demonstrates the suitability of CPA in generating features required for anomaly detection, compared to PRA which is dedicated to the task of link prediction. 

\pagebreak
CPA achieves an increase of up to 0.07 and 0.08 in precision and recall, respectively, and a decrease in the run time of up to 69 minutes in comparison to PRA. We obtained similar results with 20\% and 30\% of triples corrupted in each of the four KGs. 

\begin{table}[t!]
  \centering
  \caption{Results obtained by SEKA for entity anomaly detection.}
  \label{tab:entityanomaly}
  \begin{tabular}{lrrr}\toprule
    KG & Avg. run time (min) & Avg. precision & Avg. recall\\ 
    \midrule
    YAGO-1 & 285 & 0.93 & 0.92 \\ 
    KBpedia & 127 & 0.92 & 0.90 \\
    Wikidata & 308 & 0.90 & 0.90 \\
    DSKG & 80 & 0.91 & 0.90\\
 \bottomrule
  \end{tabular}
\end{table}

We also observe the performance of our approach in
detecting anomalous entities.
Here we consider every fact associated with an entity, including facts 
with literals ($s \in \mathbf{I}_e$ and $o \in \mathbf{L}$). While an entity can be 
anomalous due to anomalies in the facts associated with it, an entity 
can also be anomalous when there are no anomalies in the facts 
associated with it. For example, \textit{dataset 410} in DSKG is anomalous 
because it is the only dataset with eleven creators, which is a rare 
characteristic in comparison to other datasets with almost five creators. Similarly, an entity 
can be anomalous if it contains too little information compared to 
other entities. In this task of anomaly detection, our focus is on 
entities, not facts. Hence, this task has a clear distinction from 
those approaches concluding an entity as anomalous if it contains at 
least one anomalous fact.

For this experiment, we only focus on literal corruption. We
corrupted the literals associated with the entities by corrupting
their data types to introduce invalid values and removing values to
introduce missing values. We considered each entity type in the KG, where
we randomly selected a hundred entities from each type for corruption.
We corrupted 50\% of the literal-based triples belonging to each selected entity
and ran anomaly detection on each entity type of each KG. Finally,
we averaged the results obtained for each entity type to obtain the results shown in Table~\ref{tab:entityanomaly}.

As per Table~\ref{tab:entityanomaly}, our approach has high
results obtained for precision and recall demonstrating its ability
to identify anomalous entities. The run time of this approach is
relatively high compared to fact anomaly detection
because SEKA constructs a set of data quality features aimed at identifying abnormal literal-based triples.

\subsection{Anomaly Detection Without Synthetic Anomalies)}

Herewith, we provide a discussion of
some of the interesting anomalies (\textit{without} any triple corruptions) in YAGO-1, KBpedia, Wikidata, and
DSKG that we identified during the manual evaluation. We manually
evaluated the top one-hundred (based on $\nu$-SVM anomaly score) abnormal
triples in these KGs as identified by SEKA.

Consider the triple \emph{(MarcelinaChissano, bornIn, Mozambique)} which
is abnormal due to the two alternative paths \emph{(hasPredecessor,
hasPredecessor)} and \emph{(hasPredecessor, hasSuccessor)}. Investigating this
anomaly further highlights the triples \emph{(MarcelinaChissano,
hasPredecessor, GraçaMachel)}, \emph{(GraçaMachel, hasPredecessor,
Mozambique)}, and \emph{(GraçaMachel, hasSuccessor, Mozambique)}. While the
predicate \emph{bornIn} is associated with a \emph{LOCATION} as the
object, the two predicates \emph{hasPredecessor} and
\emph{hasSuccessor} are also associated with the same object which has
led to this abnormality as the rest of the triples labeled as normal
holds the relationships \emph{hasPredecessor} and \emph{hasSuccessor}
with an object of type \emph{PERSON}.


The entities \emph{Romeo (rapper)}, \emph{Roy Jones Jr.}, and
\emph{Josh Gracin} have the predicates \emph{bornIn} and
\emph{originatesFrom} with the same object. This conflicting usage of
the predicate is because these entities are musicians, and have issued
albums under their names, thus considering the composer and album as one entity. As another example,
consider the triples \emph{(MarlaMaples, hasChild, DonaldTrump)} and \emph{(MarlaMaples, isMarriedTo, DonaldTrump)} which are not abnormal independently,
but contradicting when considered as a pair.

The majority of KBpedia's anomalies are content-based as opposed to
structural anomalies. For example, the triples \emph{(NullSQL, hasLabel,
"")}, \emph{(-n-a, hasDefinition, "")}, and \emph{(-n-a, hasLabel, "")} have an empty object. Furthermore, the following triples \emph{(12-Dimethoxy-benzene, hasLabel, "Veratrol")} and \emph{(12-Dimethoxybenzene, hasLabel, "Veratrole")} provide redundant information due to minor variations in the spellings. 

In DSKG, most of the datasets
have a single website to provide more information about the dataset, known
as the landing page. The entity \emph{Dataset/1} in DSKG, has three landing
pages. Even though the three triples related to this dataset have no
anomaly in nature, they were identified as abnormal triples since this is
the only dataset with three landing pages out of which, one is a broken
link. 

\pagebreak
Similar to KBpedia, Wikidata does not have many structural anomalies. Most
of the anomalies are associated with the literals. The structural
anomalies are related to entities rich in information (triples) as
opposed to the majority with a limited amount of information. For
example, the entity \emph{L10551} has ten triples associated with it,
whereas other entities have at most two triples.

While these triples are not abnormal, the scenario of one entity having many facts
compared to the rest is anomalous. In Wikidata, literal-based anomalies
are mostly due to the absence of values, or the presence of a numerical value when
a string is required. For example, the entity \emph{L158675-F1} has an empty value for its label.

\subsection{Anomaly Detection to Compliment KG Completion}

\begin{table}[t!]
\caption[Results obtained by SEKA for KGC on the three benchmark KGs.]{Results obtained by SEKA for KGC on the three benchmark KGs. The best results are shown in bold. The acronym \textit{AD} denotes anomaly detection.}
\begin{center}
\scalebox{1.0}{
\begin{tabular}{lllrrr}
    \toprule
    Model & Measure & Particular & FB15K & FB15K-237 & WN18\\
    \hline
    \multirow{6}{*}{TransE} & \multirow{3}{*}{Hits@10} & Original & 0.75 & 0.47 & 0.94\\
    & & Random & 0.74 & 0.46 & 0.92 \\
    & & \textbf{AD} & \textbf{0.76} & \textbf{0.56} & \textbf{0.97} \\
    & \multirow{3}{*}{MRR} & Original & 0.50 & 0.29 & 0.86 \\
    & & Random & 0.42 & 0.29 & 0.86 \\
    & & \textbf{AD} & \textbf{0.60} & \textbf{0.40}  & \textbf{0.88} \\
    \hline
    
    \multirow{6}{*}{RotatE} & \multirow{3}{*}{Hits@10}  & Original & 0.88 & 0.53 & 0.96\\
    & & Random & 0.62 & 0.52 & 0.95 \\
    & & \textbf{AD} & \textbf{0.91} & \textbf{0.66} & \textbf{0.97} \\
    & \multirow{3}{*}{MRR} & Original & 0.80 & 0.34 & 0.95 \\
    & & Random & 0.79 & 0.32 & 0.94 \\
    & & \textbf{AD} & \textbf{0.81} & \textbf{0.35} & \textbf{0.97} \\
    \hline
    
    \multirow{6}{*}{SimplE} & \multirow{3}{*}{Hits@10}  & Original & 0.92 & 0.88 & 0.95\\
    & & Random & 0.91 & 0.87 & 0.95 \\
    & & \textbf{AD} & \textbf{0.93} & \textbf{0.90} & \textbf{0.97} \\
    & \multirow{3}{*}{MRR} & Original & 0.90 & 0.78 & 0.95 \\
    & & Random & 0.88 & 0.77 & 0.94 \\
    & & \textbf{AD} & \textbf{0.92} & \textbf{0.80} & \textbf{0.97} \\
    \bottomrule
\end{tabular}}
\end{center}
\label{tab:benchmarkresults}
\end{table}

We perform KGC on the three link prediction KGs FB15K\footnote{\url{https://docs.dgl.ai/en/latest/generated/dgl.data.FB15kDataset.html}}, FB15K-237\footnote{\url{https://docs.dgl.ai/en/latest/generated/dgl.data.FB15k237Dataset.html}}, and WN18\footnote{\url{https://docs.dgl.ai/en/latest/generated/dgl.data.WN18Dataset.html}}. These KGs have been introduced for the translation of embeddings in modelling multi-relational data. We use these commonly used KGs in this problem as they are relatively smaller than the KGs we used for comparative evaluation. We
present the results obtained for three of the most widely used
embedding models TransE~\cite{bordes2013transe}, RotatE~\cite{sun2019rotate}, and SimplE~\cite{kazemi2018simple} on
Table~\ref{tab:benchmarkresults}. To evaluate performance, we use Hits@10 and MRR metrics~\cite{chen2020kgc}.

As per Table~\ref{tab:benchmarkresults}, we provide results under
three particulars. We performed KGC on the original KGs without any
changes, which is given as \emph{Original}. Next, we randomly removed
some of the triples from the KGs and performed KGC. These results are
provided as \emph{Random}. 

Finally, we performed anomaly detection on
the KGs, removed the abnormal triples, and then conducted KGC. These
results are available as \emph{AD}. From the results reported in Table~\ref{tab:benchmarkresults}, we can
note that the best results for all three KGs are from the scenario of
conducting anomaly detection. Subjecting a cleansed KG for KGC after performing anomaly detection as opposed to performing KGC without any cleaning has an increase in Hits@10 of up to 0.13 and an increase in MRR of up to 0.11. Hence, it is evident that performing
anomaly detection on a KG can improve the quality of KGC.

\pagebreak
As Table~\ref{tab:runtimecomp} shows, Wikidata has the highest run time
due to it being the largest KG out of all the KGs under consideration. Also, this is the KG with the highest number of literals. Feature generation for literal-based triples is time-consuming compared to the features we generate for entity-based triples, as literal-based triples include features that calculate string similarity, and so on. Even though DSKG is the smallest KG in the cohort, it has a higher run
time than FB15K-237 due to the high number of literals it has, which
we use for entity-based anomaly detection. Hence, the total run time
of our approach is primarily dependent on the feature construction
step. As an increase in the number of nodes and edges, and features
does not cause an exponential increase in run time, this provides an
indication of the scalability of our approach and its suitability on
large KGs.

\begin{table}[t]
\caption[Run time information of SEKA.]{Total run time of SEKA for the four original KGs during fact anomaly detection.}
\label{tab:runtimecomp}
  \centering
  \begin{small}
  \begin{tabular}{lc}
  \toprule
    KG & Total run time (minutes)\\ 
    \midrule
    YAGO-1 & 121 \\ 
    KBpedia & 48 \\
    Wikidata & 232 \\
    DSKG & 72 \\
    FB15K & 92\\
    FB15K-237 & 56\\
    WN18 & 40\\
 \bottomrule
  \end{tabular}
  \end{small}
\end{table}

\subsection{Evaluation of TAXO}
\label{subsec:experimentstaxo}

Using YAGO-1\footref{fn4},
KBpedia\footref{fn5},
Wikidata\footref{fn6}, and DSKG\footref{fn7} we evaluate the extent to which we can discover the anomaly types identified in Section~\ref{sec:taxo} without the adoption of a complex technique such as an anomaly detection approach. Hence, we propose a rule-based approach, where we define the rules of it in Table~\ref{tab:rulebasedapproach}. Next, we compare the output of SEKA with the rule-based approach to determine the coverage of each of the approaches. 
\newline

\noindent\textbf{A Rule-based Approach for Anomaly Detection }

Herewith, we first provide the preliminaries, where we define the notations we use to present the rules, and then provide the set of rules in Table~\ref{tab:rulebasedapproach} that we use to detect anomalies as classified by TAXO in Section~\ref{sec:taxo}. With reference to the preliminaries in Section~\ref{sec:preliminaries}, we represent a KG as $G=(V,E)$, and its storage file as $G_s$. We use the notations $s$, $p$, and $o$ to represent a triple $F$, where $F = (s, p, o)$, $o \in \mathbf{L}$, or $o \in \mathbf{I}_e$. Similarly, $F_1 = (s_1, p_1, o_1)$ and $F_2 = (s_2, p_2, o_2$). 

\begin{table*}[t!]
    \centering
    \caption[Rule-based approach for anomaly detection.]{Rules to identify anomalies as specified by TAXO in Section~\ref{sec:taxo}. The exclamation mark (!) in the following table denotes the NOT operator.}
    \label{tab:rulebasedapproach}
    \begin{tabular}{p{1.5in}p{3.65in}}
    \toprule
    Anomaly type & Rule\\
    \midrule
    Missingness & IF [($s$ == null) OR ($p$ == null) OR ($o$ == null)] THEN $F$ has a missing element.\\
    \\[-1.0em]
    
    Incorrectness & No rule can be defined for entity-based triples.\\
    \\[-1.0em]
    
    & IF [($o \in \mathbf{L}$) AND (dataType ($o$) $\neq$ type ($p$)]
    THEN $F$ has an incorrect literal.\\
    \\[-1.0em]
    
    Partial incorrectness & No rule can be defined.\\
    \\[-1.0em]
    
    Inconsistency & IF [!coOccur ($p$, $s$) OR !coOccur ($p$, $o$)] THEN $F$ is inconsistent.\\
     \\[-1.0em]
     
    Entity ambiguity & IF ($s == 0$) THEN $F$ has entity ambiguity.\\
    \\[-1.0em]
     & IF [($s_1$ == $s_2$) AND ($o_1$ == $o_2$) AND (($type (s_1) \neq type (s_2)$) OR ($type (o_1) \neq type (o_2)$))] THEN $F_1, F_2$ have entity ambiguity.\\
     & IF [(($s_1$ == $o_2$) AND ($type (s_1) \neq type (o_2)$)) OR (($s_2$ == $o_1$) AND ($type (s_2) \neq type (o_1)$))] THEN $F_1, F_2$ have entity ambiguity.\\

    \\[-1.0em]
    Predicate ambiguity & IF [($s_1$ == $s_2$) AND ($p_1$ == $p_2$) AND ($o_1$ is\_a $o_2$)] THEN $F_1$, $F_2$ have a predicate ambiguity.\\
    \\[-1.0em]
    
    Contradictions & IF [($s_1$ == $s_2$) AND ($o_1$ == $o_2$) AND !coOccur ($p_1$, $p_2$)] THEN $F_1, F_2$ are contradicting.\\
    \\[-1.0em]
    
    Rare entity & FOREACH $v \in V$; IF (triplesCount ($v$) $\leq 1$), THEN $v$ is rare. \\
    \\[-1.0em]
    
    Prolific entity & FOREACH $v \in V$; IF [triplesCount ($v$) $\ge avgTripleCount(V)$], THEN $v$ is prolific. \\
    \\[-1.0em]
    
    Redandancies & IF [($s_1$ == $s_2$) AND ($p_1$ == $p_2$) AND (similarity($o_1$,$o_2$) $\ge 0.8$)] THEN $F_1, F_2$ are redundant. We consider a similarity score of 0.8 to only consider the most similar strings.\\
    \\[-1.0em]
    
    Duplicates & IF [($s_1$ == $s_2$) AND ($p_1$ == $p_2$) AND ($o_1$ == $o_2$)] THEN $F_1, F_2$ are duplicates.\\
    \bottomrule
    \end{tabular}
    
\end{table*}

We use ENTGENE introduced in Section~\ref{subsec:entgene} to generate the type information of entities, and TRIC~\cite{senaratne2023tric} to infer predicate usage. While we are able to generate rules for most of the anomaly types as listed in TAXO in Section~\ref{sec:taxo}, we could not generate rules to identify incorrect triples, as determining incorrectness (when there is no semantic or structural anomaly) requires input from domain experts, or from an external resource. However, we generated a rule to identify triples with an invalid literal usage (such as the triple \textit{(DJShadow, created, "album")}), where we categorize such triples as incorrect. With the use of TRIC~\cite{senaratne2023tric}, we get data type information of literals most commonly used with a particular predicate within the KG.

In Table~\ref{tab:sekarules}, we provide the results of our experimental evaluation. Pertaining to TAXO, Table~\ref{tab:sekarules} shows the counts of each anomaly type identified from the KGs by SEKA, where we manually verified and classified the top 100 abnormal triples (based on $\nu$-SVM~\cite{scholkopf2001svm} score) for each of the anomaly types in TAXO. As per the results in Table~\ref{tab:sekarules}, even though some anomaly types show that SEKA has not detected any anomalous triples from that category, this means that triples reflecting that particular anomaly do not belong to the top 100 anomalies based on $\nu$-SVM.
Furthermore, Table~\ref{tab:sekarules} also shows how many of the anomalies out of those identified by SEKA (within its top 100 anomalies) can be identified by the rule-based approach.

\begin{table*}[t!]
    \centering
    \caption[Comparison of SEKA and the rule-based approach.]{Number of anomalies in each KG pertaining to TAXO, as identified by SEKA and the rule-based approach within the top 100 anomalies identified by SEKA based on $\nu$-SVM score. We use 'S' to denote SEKA, and 'R' to denote the rule-based approach.}
    \label{tab:sekarules}
    \scalebox{0.8}{
    \begin{tabular}{llcccccccc}
    \toprule
     \multirow{2}{*}{Triple type} & \multirow{2}{*}{Anomaly name} & \multicolumn{2}{c}{YAGO-1} & \multicolumn{2}{c}{Wikidata} & \multicolumn{2}{c}{DSKG} & \multicolumn{2}{c}{KBpedia} \\
     \cline{3-10}
     & & S & R & S & R & S & R & S & R\\
    \midrule
    \multirow{12}{*}{Entity-based} & Missing subject & 2 & 2 & 0 & 0 & 0 & 0 & 0 & 0\\
     & Missing predicate & 2 & 2 & 0 & 0 & 0 & 0 & 0 & 0 \\
     & Missing object & 0 & 0 & 0 & 0 & 0 & 0 & 1 & 1 \\
     & Incorrect link & 5 & 0 & 0 & 0 & 0 & 0 & 0 & 0 \\
     & Invalid predicate & 12 & 0 & 6 &  0 & 0 & 0 & 1 &  0 \\
     & Entity ambiguity & 10  & 4 & 0 &  0 & 0  & 0 & 8 &  7 \\
     & Predicate ambiguity & 8  & 8 & 13 &  2 & 0 &  0 & 9  & 6\\
     & Contradicting facts & 9 &  0 & 0  & 0 & 0  & 0 & 4  & 0\\
     & Rare entity & 6  & 6 & 12 &  12 & 61 &  61 & 12  & 12 \\
     & Prolific entity & 2  & 2 & 15 &  15 & 12  & 12 & 11 &  11 \\
     & Redundant facts & 6  & 6 & 0 &  0 & 0 &  0 & 9 &  9\\
     & Duplicate facts & 2 &  2 & 1  & 1 & 14  & 14 & 0 &  0\\
    
    \cline{1-10}
    Entity \& literal-based & Mixed triples & 2 &  0 & 4 &  0 & 0 &  0 & 0  & 0\\
    \cline{1-10}
    \multirow{8}{*}{Literal-based} & Missing subject & 1 &  1 & 0 &  0 & 0  & 0 & 0  & 0\\
     & Missing predicate & 3 &  3 & 0 &  0 & 0 &  0 & 0 &  0\\
     & Missing object & 6 &  6 & 12  & 12 & 0 &  0 & 14  & 14\\
     & Incorrect literal & 5 &  5 & 13  & 13 & 0  & 0 & 11  & 11\\
     & Partially correct literal & 8 &  0 & 0 &  0 & 0 &  0 & 0  & 0\\
     & Invalid predicate & 5  & 5 & 0 &  0 & 0  & 0 & 0 &  0\\
     & Redundant literals & 6  & 6 & 12  & 12 & 0  & 0 & 18  & 18\\
     & Duplicates & 0 &  0 & 12 &  12 & 13 &  13 & 2 &  2\\
    \bottomrule
    \end{tabular}}
\end{table*}

As can be seen from the results in Tables~\ref{tab:sekarules}, SEKA has a wide variety of anomalies detected. While SEKA has detected more invalid predicates and triples with entity ambiguity within its top 100 for YAGO-1 (compared to the other three KGs), SEKA has detected more rare and prolific entities from the other three KGs Wikidata, DSKG, and KBpedia (compared to YAGO-1. On the other hand, SEKA has also detected triples with missing objects and incorrect literals within its top 100 results for all four KGs, respectively. While SEKA has detected triples with data quality errors, the top 100 anomalies mostly contain triples with semantic errors and entities with anomalies.

Similarly, the results in Table~\ref{tab:sekarules} show how well the rule-based approach performed in detecting those anomalies identified by SEKA within its top 100. We manually verified the output of this approach and performed a manual classification at instances when the rules made an incorrect classification. While the rules-based approach is capable of identifying those errors related to data quality aspects, it has missed most of the semantic-related and entity-related anomalies. As rules tend to evaluate each triple individually, it is common for such an approach to miss those anomalies related to inter-related triples.

As the types of anomalies detected by SEKA are comparatively higher than that of the rule-based approach, we can conclude that having an anomaly detection approach is beneficial over an approach that individually assesses triples in a KG. That is, SEKA has more coverage than the set of pre-defined rules. Similar to the rule-based approach, SEKA cannot detect incorrect triples as it does not depend on external sources. Hence, it is evident that we cannot always pre-define an anomaly, as they do not have a common pattern to be guessed. However, SEKA can detect incorrect triples as they emerge due to a relationship they hold with another triple that leads to a different anomaly such as contradictions. It is also important to note that anomaly detection does not always require a complex solution, as detection of certain anomalies such as rare/frequent occurrences, duplicates and redundancies is easy. However, a complex solution can discover a super-set of anomalies.

\pagebreak
\section{Conclusion and Future Work}
In this paper, we introduced two techniques SEKA (\underline{Se}eking \underline{K}nowledge Graph \underline{A}nomalies) and TAXO (\underline{Taxo}nomy of anomaly types in KGs), for anomaly detection and classification in Knowledge Graph (KG). While SEKA is an approach to detect anomalous triples and entities in a KG in an unsupervised manner, TAXO is a taxonomy (classification) of possible anomaly types occurring in KGs. 

SEKA can identify anomalies related to both the structure and content of the KG, it is independent of external resources and has the ability to identify a multitude of anomalies without depending on any external resource. A triple is considered anomalous if it is semantically incorrect, contradicting, has incorrect entity type information, provides redundant information, or has an invalid/missing literal. 

Having detected anomalies, TAXO categorizes anomalies based on the triple type the anomaly occurs as entity-based, literal-based, and both entity and literal-based. Afterwards, the categorization further delves into the number of triples required to form each anomaly, as single or multiple triples. TAXO elaborates on the anomaly types by providing an example for each identified anomaly type and a possible approach for rectifying the anomaly. 

We evaluated both approaches using the four real-world KGs YAGO-1, DSKG, Wikidata, and KBpedia. The results of the experimental evaluation show how SEKA outperforms its baselines for all four KGs with substantially reduced run times, and higher precision and recall values. Furthermore, the experimental results show how KG refinement via anomaly detection can improve the quality of Knowledge Graph Completion (KGC). Furthermore, as per the experimental results of the performance of SEKA and the rule-based approach for anomaly detection, it is evident that an anomaly detection approach has more coverage over a rule-based approach that functions on a set of pre-configured rules.

As future work, we aim to focus our research direction towards anomaly detection in dynamic KGs~\cite{vsiljak2008dynamicgraphs}. Following our line of work, we wish to introduce an approach that can incorporate the time dimension into features, such that we generate semantically meaningful temporal features.

\clearpage

\bibliographystyle{elsarticle-harv} 
\bibliography{bibliography}

\end{document}